\documentclass[letterpaper, 10 pt, conference]{ieeeconf}  

\IEEEoverridecommandlockouts                              

\makeatletter
\def\endthebibliography{%
	\def\@noitemerr{\@latex@warning{Empty `thebibliography' environment}}%
	\endlist
}
\makeatother

\usepackage{graphics} 
\usepackage{epsfig} 
\usepackage{mathptmx} 
\usepackage{times} 
\usepackage{amsmath} 
\usepackage{amssymb}  
\usepackage{siunitx}
\usepackage{verbatim}
\usepackage{flushend}
\usepackage{mathrsfs}
\usepackage{booktabs}
\usepackage{multirow}
\usepackage{upgreek}

\usepackage[ruled,vlined]{algorithm2e} 

\title{\LARGE \bf  Intensity Image-based LiDAR Fiducial Marker System}

\author{Yibo Liu, Hunter Schofield, Jinjun Shan
\thanks{This work was supported in part by NSERC Alliance Program under Grant ALLRP 555847-20, and in part by Mitacs Accelerate Program under Grant IT26108. The authors are with Department of Earth and Space Science and Engineering, York University, Toronto, Ontario M3J 1P3, Canada 
        {\tt\footnotesize \{yorklyb,hunterls,jjshan\}@yorku.ca}}%
}

\begin{document}

\maketitle
\thispagestyle{empty}
\pagestyle{empty}

\begin{abstract}
The fiducial marker system for LiDAR is crucial for the robotic application but it is still rare to date. In this paper, an Intensity Image-based LiDAR Fiducial Marker (IILFM) system is developed. This system only requires an unstructured point cloud with intensity as the input and it has no restriction on marker placement and shape. A marker detection method that locates the predefined 3D fiducials in the point cloud through the intensity image is introduced. Then, an approach that utilizes the detected 3D fiducials to estimate the LiDAR 6-DOF pose that describes the transmission from the world coordinate system to the LiDAR coordinate system is developed. Moreover, all these processes run in real-time (approx 40 Hz on Livox Mid-40 and approx 143 Hz on VLP-16). Qualitative and quantitative experiments are conducted to demonstrate that the proposed system has similar convenience and accuracy as the conventional visual fiducial marker system. 
The codes and results are available at: https://github.com/York-SDCNLab/IILFM.
\end{abstract}

\section{INTRODUCTION}
Cameras and LiDARs are two types of vital sensors found in autonomous driving and robotic applications \cite{kt,barfoot}. Research on the Visual Fiducial Marker (VFM) system, the fiducial marker system developed for the camera, has a long history \cite{art} and lavish experience and research achievements have been accumulated \cite{wang,olson,ap3,aruco,cctag}. The VFM provides the environment with controllable artificial features such that the feature extraction and matching become simpler and more reliable. The VFM systems have been applied in augmented reality \cite{ar}, human-robot interaction \cite{wang,olson}, multi-sensor calibration \cite{kalibr,lt2} and Simultaneous Localization and Mapping (SLAM) \cite{munoz,munoz2019}.

Development of a LiDAR fiducial marker system also plays a key role in LiDAR applications since feature extraction and matching are fundamental requirements in LiDAR applications \cite{lt2,lio,loam}. Moreover, as introduced in \cite{lt}, the SLAM framework that utilizes the artificial features provided by the LiDAR fiducial marker becomes feasible, which will be an interesting new robotic application. In contrast with the VFM system, the fiducial marker system for LiDAR is rare. To the best of our knowledge, up to date the only fiducial marker system for LiDAR that has similar functionality as the VFM is LiDARTag \cite{lt}. However, LiDARTag has a spatial restriction on marker placement on account of the fact that it employs conventional single-linkage clustering algorithm to find the clustering of the marker in the point cloud \cite{lt}. Consequently, to satisfy the spatial restriction on marker placement, it is required to add an extra 3D object to the environment when adding a LiDARTag. This negatively affects the spatial environment.
\vspace{-0.05in}
\begin{figure}[htpb]
	\centering
	\includegraphics[width=2.5in]{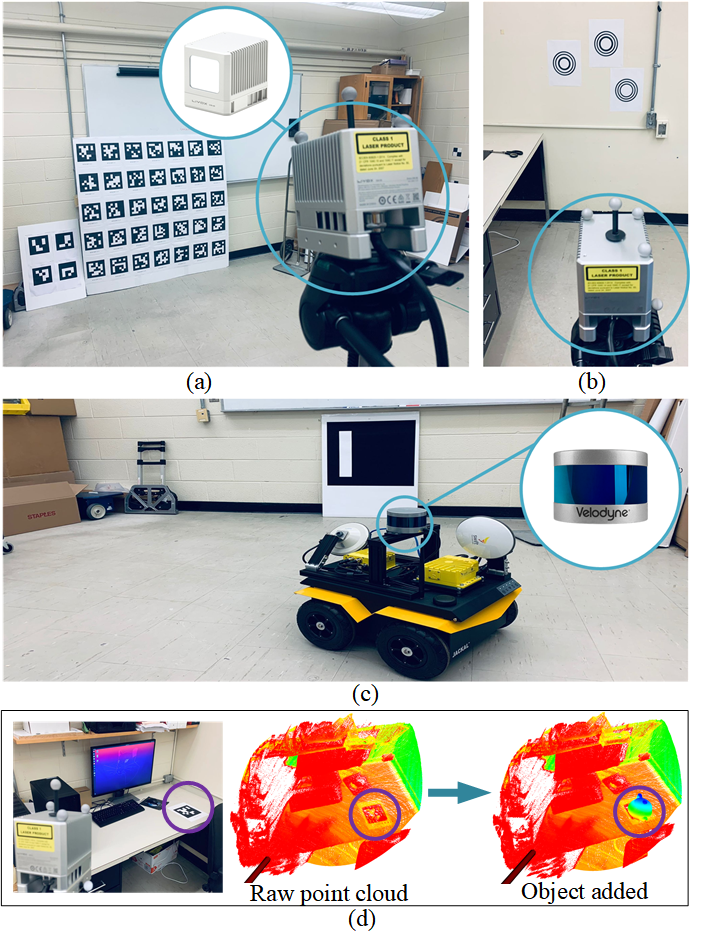}
	\vspace{-0.1in}
	\caption{The proposed system has ample flexibility. Different visual fiducial marker systems (Apriltag \cite{ap3}, ArUco \cite{aruco} and CCTag \cite{cctag}, etc.) can be easily embedded; There is no restriction on marker placement; The system is applicable for both solid-state LiDAR (a, b) and mechanical LiDAR (c). An augmented reality demo using the proposed system is shown in (d): the teapot point cloud is transmitted to the location of the marker in the LiDAR's point cloud based on the pose provided by the IILFM system.} \label{flex}
\end{figure}
\begin{figure}[thpb]
	\centering
	\includegraphics[width=2.8in]{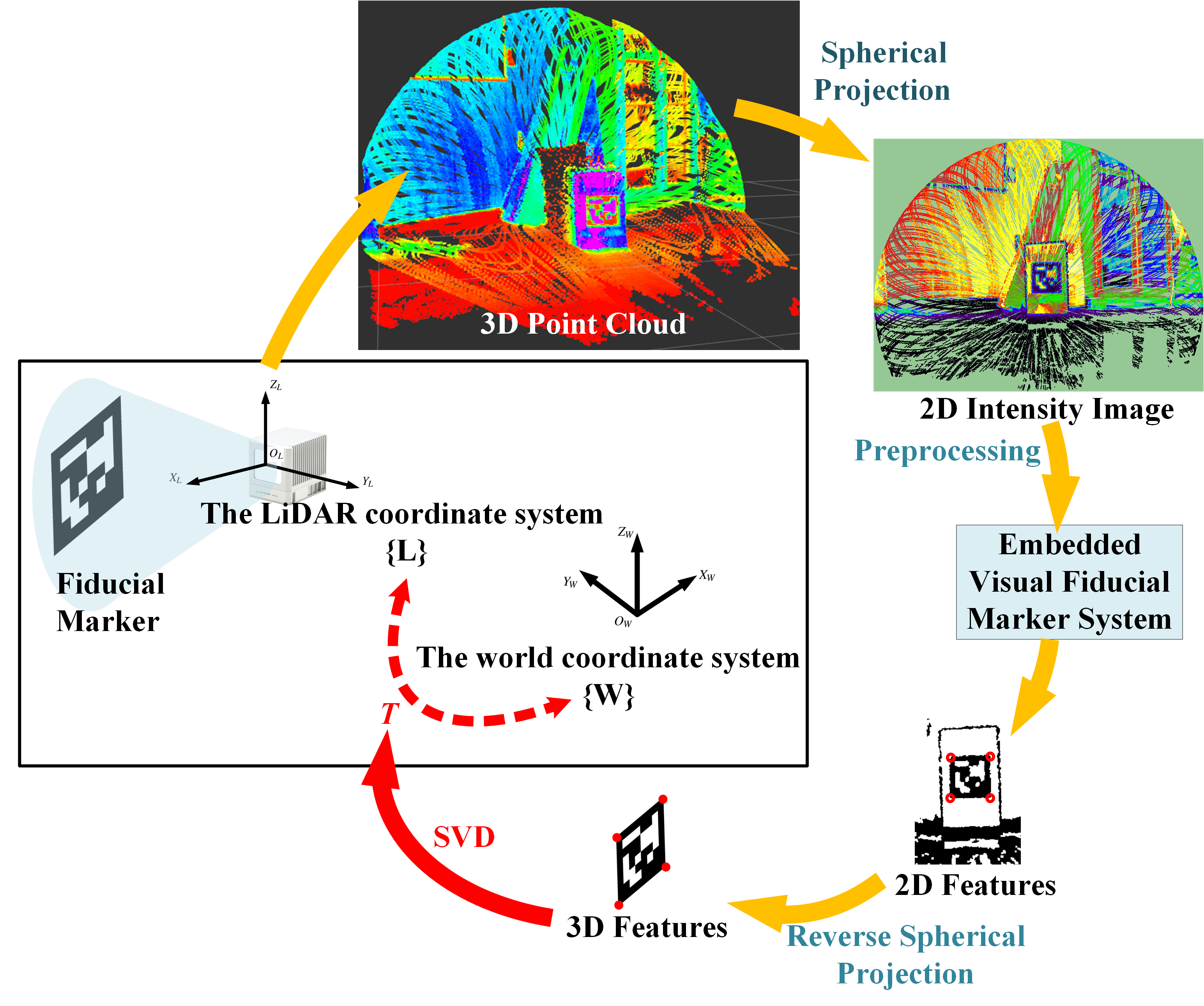}
	\vspace{-0.1in}	
	\caption{An overview of the proposed Intensity Image-based LiDAR Fiducial Marker detection and pose estimation pipeline. The 3D point cloud is first transformed into an intensity image by spherical projection. Then the embedded VFM detector is adopted on the preprocessed intensity image. The detected 2D features are then projected back to 3D space. Finally, the pose of the LiDAR is estimated by optimally aligning the point set of the predefined 3D features and the point set of the detected 3D features.} \label{overview}
\end{figure}

Reviewing the VFM systems, it is found that there are no restrictions on marker placement \cite{ap3,aruco,cctag}. Namely, the marker can be placed anywhere without affecting the environment in terms of space. According to \cite{lt}, the abandonment of this free-placement virtue is owed to the gap between the structured image and unstructured point cloud. Nevertheless, this gap is not insurmountable. In particular, there is a notable hot trend in LiDAR-based 3D object detection/segmentation  \cite{rangenet,lasernet}: Neural Networks that are originally developed for 2D object detection/segmentation can be utilized to detect/segment the objects in the range/intensity image(s) of the 3D LiDAR point cloud. This indicates that the range/intensity image(s) generated from the LiDAR point cloud can be a pathway to transfer the research accomplishments on the 2D image to the 3D point cloud. Following this inspiration and also considering that the black-and-white marker is explicitly visible in the point cloud rendered by the intensity (See Fig.~\ref{overview}), we propose an Intensity Image-based LiDAR Fiducial Marker (IILFM) system in this paper. The ample flexibility of the IILFM system is illustrated in Fig.~\ref{flex} and the overview pipeline of the system is shown in Fig.~\ref{overview}. The contributions of this paper are:
\begin{itemize}
\item The development of a novel fiducial marker system for the LiDAR, the IILFM system. Unlike LiDARTag \cite{lt} which requires extra 3D objects to be added to the environment, the application of the IILFM is as convenient as the VFM systems \cite{ap3,aruco}. 
\item The proposal of a novel marker detection method to detect the predefined 3D features through the intensity image. Thanks to this, the VFM systems proposed in the past, present, and even future can be easily embedded into the IILFM system.
\item The introduction of a pose estimation approach for the LiDAR via the proposed IILFM, which has similar accuracy as the VFM-based pose estimation for the camera. In addition, dissimilar to the VFM system, the proposed pose estimation is free of the rotation ambiguity problem \cite{ippe,yibo}.
\end{itemize}

The paper is organized as follows. Section~\ref{stwo} introduces the related work. Section~\ref{sthree} presents the pipeline for detecting 3D fiducials in the point cloud through the intensity image. Section~\ref{pose} illustrates the method to estimate the LiDAR's pose with respect to the world coordinate system. Section~\ref{exp} provides the qualitative and quantitative experimental evaluations regarding marker detection and pose estimation. Section~\ref{con} gives the conclusion and future work.

\section{Related Work\label{stwo}}
The previous research on the VFM system mainly focuses on the following four aspects: 

(1) A higher detection rate. Take the process to extract the line segments in the image as the example, ARToolkit \cite{art} adopts naive thresholding; the first generation of Apriltag \cite{olson} employs image gradients to detect high-contrast edges, which has a similar effect as binarization; and the second generation of Apriltag \cite{wang} uses adaptive thresholding.  The objective of these improvements is to boost the detection of line segments, candidate quads \cite{ap3,aruco}, or circles \cite{cctag}, thus the marker detection rate under varying ambient light is improved. 

(2) A lower false positive rate. The coding/decoding systems of the latest VFM systems \cite{ap3,aruco} are upgraded with regard to the old generations \cite{wang,olson},  such that the identification of a candidate marker is more reliable and the false positive cases are reduced when decoding. 

(3) A lower computational time. Again, through the amelioration of methods for graphic segmentation and algorithms adopted in coding/decoding systems, the VFM systems are accelerated. For instance, the speed of the third generation of Apirltag \cite{ap3} is almost 5 times faster than that of the second generation \cite{wang}. 

(4) Resolving the rotation ambiguity problem. The rotational ambiguity means a planar marker could project onto the same pixels from two different poses when the perspective effect is weak \cite{ippe}. Much research \cite{yibo,jin,ch2020} has been conducted to resolve the duo solutions problem. Unlike the first three aspects, the last aspect is a remaining fundamental challenge.

The VFM system is embedded in the proposed system, such that the achievements mentioned in aspects (1)-(3) are inherited in the proposed one. Furthermore, as shown in Section~\ref{pose}, the proposed system is free from the problem introduced in aspect (4).

Although LiDARTag \cite{lt} also borrows the dictionary-based coding idea from the Apriltag system \cite{ap3}, most of the accumulations mentioned above are not preserved since LiDARTag follows the conventional approaches to process the 3D point cloud, which brings forward two inconveniences. Firstly, to utilize the single-linkage clustering algorithm, it is needed to guarantee that the representation of the marker in the point cloud meets a spatial requirement: the marker must have $t\sqrt{2}/ 4$ clearance around it \cite{lt} where $t$ represents the marker's size. If the marker is attached to a wall or a box, it is required to make sure $\tau>t\sqrt{2}/ 4$ where $\tau$ is the thickness of the marker's 3D object. Therefore, consider the case that 10 LiDAR Tags are placed in an indoor environment. Such a situation will result in a room crowded with tripods and large markers. This apparently wrecks the environment. The second inconvenience is that the algorithms to process the point cloud and point clustering are packaged in the implementation of LiDARTag which assumes that the marker is square and the pattern belongs to Apriltag \cite{ap3}. Suppose that it is desired to adopt a non-square marker system, such as CCTag \cite{cctag}, then it is required to reprogram the LiDARTag system.

\section{Marker Detection\label{sthree}}
\subsection{Generation of the Intensity Image} \label{sa} 
The generation of an intensity image from a given unstructured point cloud can be summarized as transferring all the 3D points in the point cloud onto a 2D image plane by spherical projection and rendering the corresponding pixels with intensity values. 
\begin{figure}[thpb]
	\centering
	\includegraphics[width=2.8in]{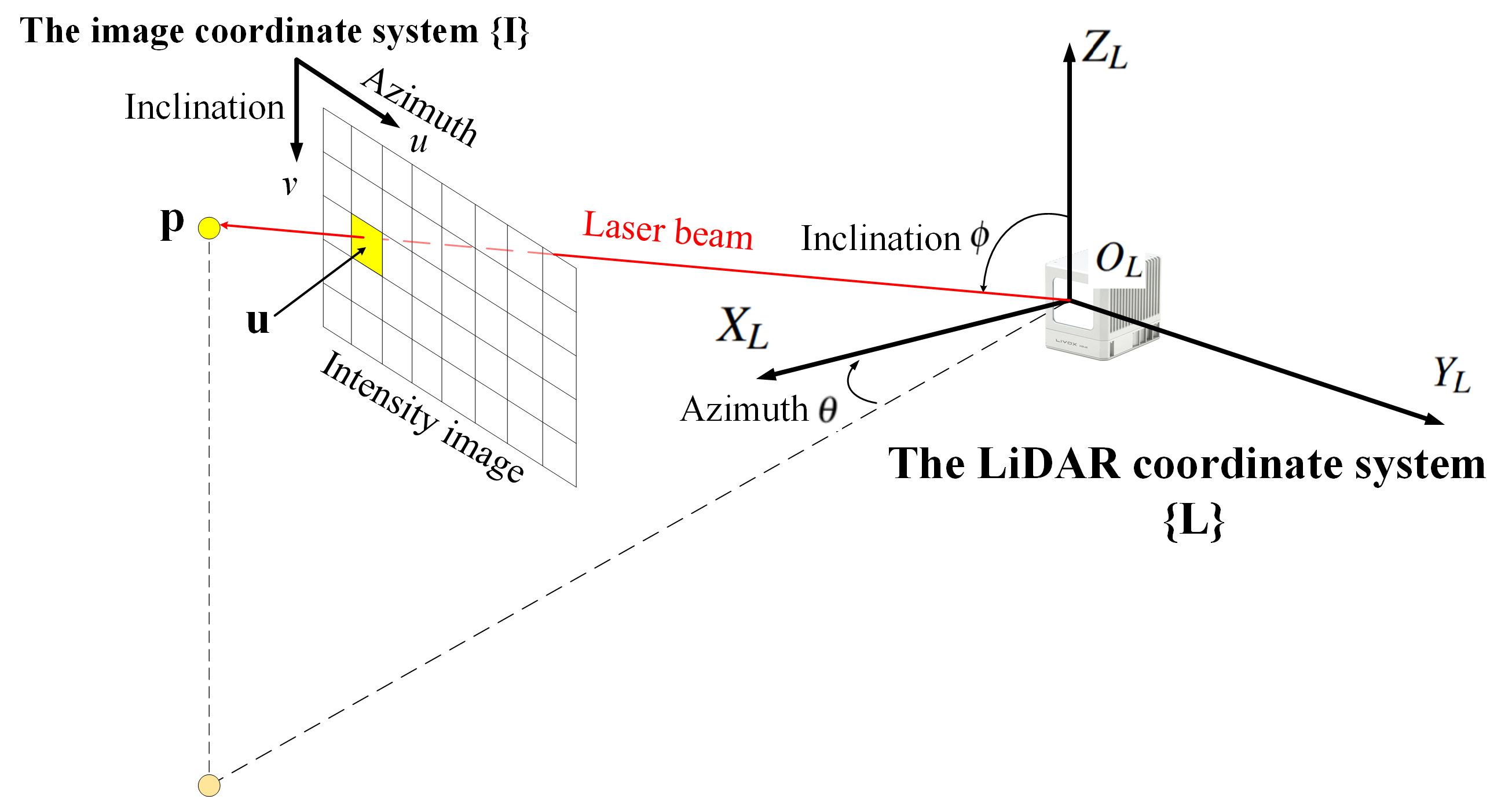}
	\caption{An illustration of the coordinate systems and notations.}
	\label{notation}
\end{figure}

Fig.~\ref{notation} shows the coordinate systems and notations. $\textbf{p}=[x_{l},y_{l},z_{l},i]^{T}$ is an observed point in the 3D point cloud, with ${[x_{l},y_{l},z_{l}]^{T}}$ denoting its Cartesian coordinates w.r.t the LiDAR coordinate system \textbf{\{L\}} and $i$ being the intensity.  $\textbf{u}$ is the projection of \textbf{p} onto the image plane, which has coordinates $[u,v]^{T}$ w.r.t the image coordinate system, \textbf{\{I\}}. Following \cite{barfoot}, the Cartesian coordinates of \textbf{p} are first transformed to spherical coordinates $[\theta,\phi,r]^{T}$:
\begin{equation}	
	\begin{aligned}
		& \theta=\arctan(\frac{y_{l}}{x_{l}})\\
		& \phi=\arctan(\frac{z_{l}}{\sqrt{x_{l}^{2}+y_{l}^2}})\\
		& r=\sqrt{x_{l}^2+y_{l}^2+z_{l}^2}
	\end{aligned}\label{pro}
\end{equation}
where $\theta$ and $\phi$ denote the azimuth and inclination, respectively. $r$ is the range from \textbf{p} to the origin of \textbf{\{L\}}.

Then, the image coordinates $[u,v]^{T}$ of  $\textbf{u}$ are given by:
\begin{equation}	
	u = \lceil \frac{\theta}{\Theta_{a}}\rfloor + u_{o},\; v = \lceil \frac{\phi}{\Theta_{i}}\rfloor + v_{o} \label{uv}
\end{equation}
where $ \lceil \;  \rfloor$ represents rounding a value to the nearest integer. $\Theta_{a}$ and $\Theta_{i}$ are the angular resolutions in $u$ (azimuth) and $v$ (inclination) directions, respectively. $u_{o}$ and $v_{o}$ are the offsets. 

Assume it is desired that the point with zero-azimuth and zero-inclination to be projected to the center of the image, then the offsets will be: $u_{o}=I_{w}/2$ and $v_{o}=I_{h}/2$, where $I_{w}$ and $I_{h}$ being the image width and height which are determined by the maximum angular width $P_{w}$ and height $P_{h}$ of the point cloud $I_{w} = \lceil \frac{P_{w}}{\Theta_{a}}\rfloor, \; I_{h} = \lceil \frac{P_{h}}{\Theta_{i}}\rfloor$. The pixel $[u,v]^{T}$ is then rendered using the corresponding intensity value $i$, and the range information $r$ is also saved for the sake of the following pose estimation. Thereafter, we step through the point cloud and repeat the above process. The pixels that are not mapped to any points will remain unobserved and are rendered by a unique predefined value. Note that if these pixels are visited later on in the marker detection process, they will not return any 3D points because they represent unobserved regions.

\noindent\textbf{\textit{Remark 1}}: As a matter of fact, the rounding in the Eq. (\ref{uv}) causes information loss if the pixels are non-integer values at the beginning. However, this is inevitable when transferring the points in the point cloud onto an image plane where the points are expressed with integer coordinates. Furthermore, this approximation is not fatal as can be seen in Section~\ref{exp}, the pose estimation precision is still neck-to-neck with the VFM system despite the information loss.

\subsection{Selections of the Angular Resolutions} \label{sb}
\begin{figure*}
	\centering
	\includegraphics[width=17cm]{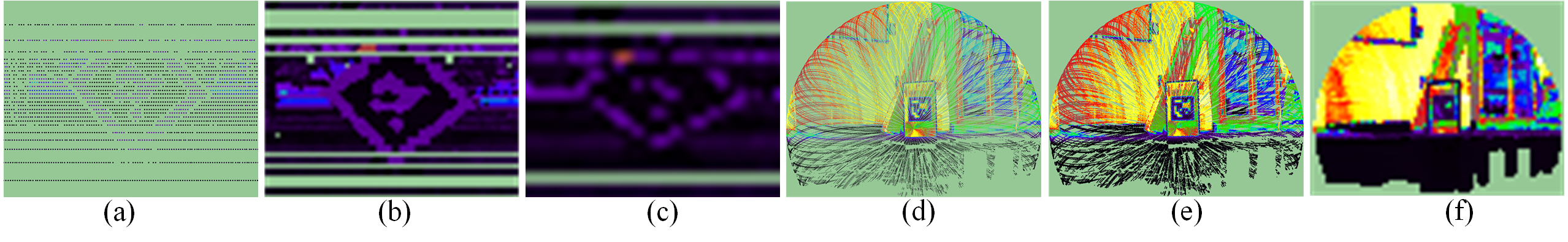}
	\vspace{-0.1in}
	\caption{The intensity images generated under different angular resolution settings.  The unobserved regions are rendered light green. (a), (b), and (c) are intensity images generated from the same point cloud given by one LiDAR scan of a Velodyne ULTRA Puck (mechanical LiDAR, $\Theta_{h}=\textbf{0.4}^{\circ}$ and $\Theta_{v}=\textbf{0.33}^{\circ}$). The LiDAR scan is extracted from the dataset provided by \cite{lt}. (d), (e), and (f) are intensity images generated from the same point cloud which is the integration of multiple LiDAR scans of a Livox Mid-40 (solid-state LiDAR,  $\Theta_{h}=\Theta_{v}=\textbf{0.05}^{\circ}$). (a), (b), and (c) are cropped to display. \textbf{(a)} $\Theta_{a}=0.05^{\circ}$ and $\Theta_{i}=0.05^{\circ}$. Raw image size = $7200\times800$.  \textbf{(b)} $\Theta_{a}=0.4^{\circ}$ and $\Theta_{i}=0.33^{\circ}$. Raw image size = $900\times121$. \textbf{(c)} $\Theta_{a}=0.6^{\circ}$ and $\Theta_{i}=1.0^{\circ}$. Raw image size = $600\times40$.  \textbf{(d)} $\Theta_{a}=\Theta_{i}=0.025^{\circ}$. Raw image size = $1538\times1178$.  \textbf{(e)} $\Theta_{a}=\Theta_{i}=0.05^{\circ}$. Raw image size = $771\times591$. \textbf{(f)} $\Theta_{a}=\Theta_{i}=0.5^{\circ}$. Raw image size = $81\times62$.} \label{newreso}
\end{figure*}

\begin{figure}[htbp]
	\centering
	\includegraphics[width=3.0in]{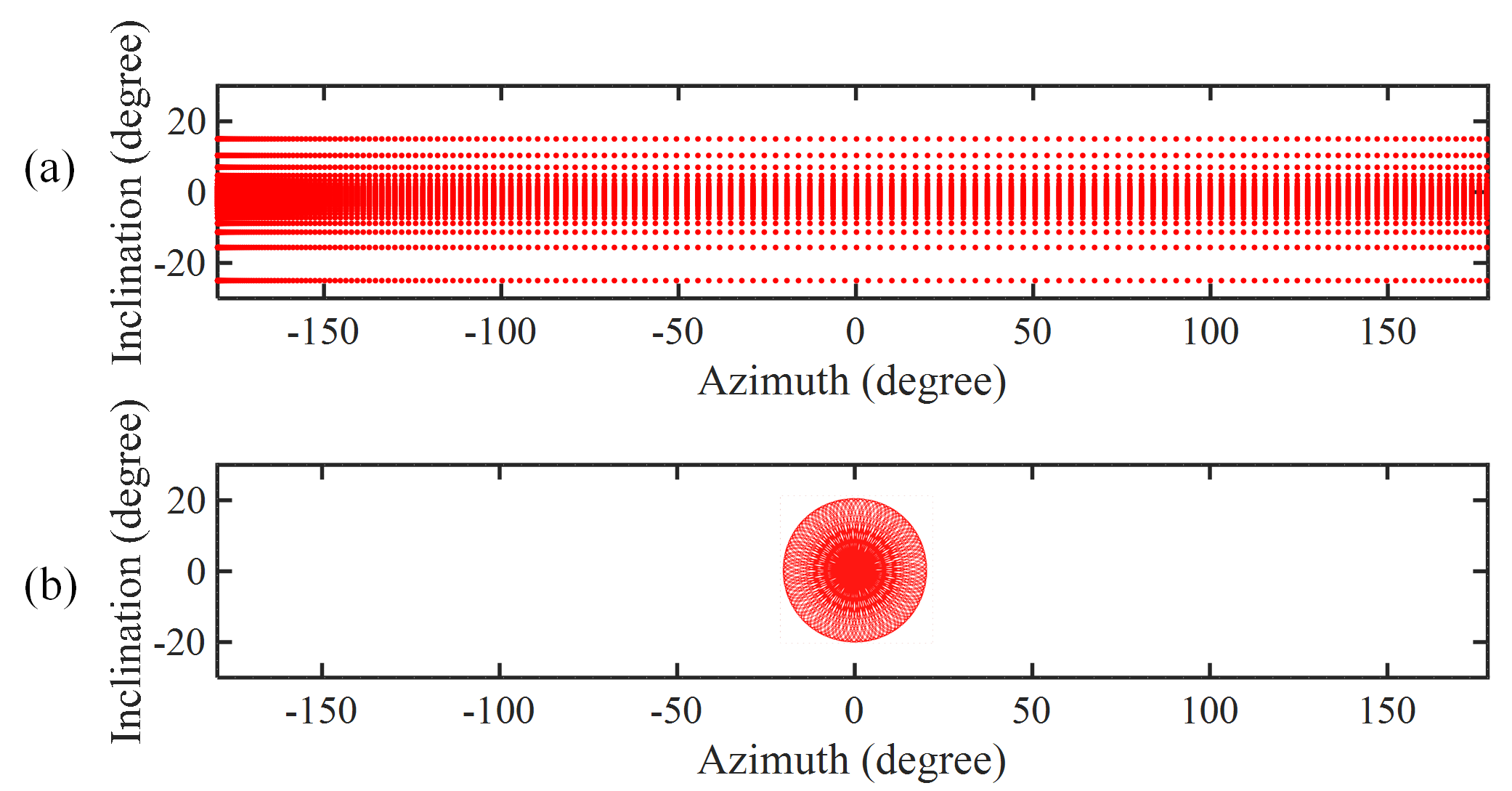}
	\caption{Sampling patterns of the mechanical LiDAR and the solid-state LiDAR. Sampling points are represented by red scatters. (a) General sampling pattern of a mechanical LiDAR expressed in the azimuth/inclination coordinate system. This is a general schematic that does not correspond to any LiDAR model. (b) Sampling pattern of the Livox Mid-40 LiDAR after one-second integration. Note that the sampling patterns vary when it comes to different solid-state LiDAR models. But generally, the unscanned regions within the valid Field of View (FoV) appear as spots.} \label{samplecom}
\end{figure}

The selections of $\Theta_{a}$ and $\Theta_{i}$ are crucial as they affect the quality of the intensity image directly. Yet, the resolution of this problem is quite simple: suppose that  $\Theta_{h}$ is the horizontal angular resolution and $\Theta_{v}$ is the vertical angular resolution of the employed LiDAR, set $\Theta_{a}=\Theta_{h}$ and $\Theta_{i}=\Theta_{v}$.  Fig.~\ref{newreso} illustrates the reasoning behind this setting. There are two objectives when generating the intensity image: 1) we want fewer unobserved regions to appear around or inside the marker's area since they do not provide any intensity or range information. Moreover, they could become image noise after the image preprocessing phase. Thus, the settings shown in Fig.~\ref{newreso} (a) and (d) cannot be used, where $\Theta_{a}<\Theta_{h}$ and $\Theta_{i}<\Theta_{v}$; 2) However, if a dense image is desired, we need to be aware that too much information will be lost if too many points overlap for the same pixels based on Eq.~(\ref{uv}). For instance, the marker's pattern disappears in Fig.~\ref{newreso} (c) and (f), where $\Theta_{a}>\Theta_{h}$ and $\Theta_{i}>\Theta_{v}$, even though they are dense images. For the sake of the two objectives, the following settings should be used: $\Theta_{a}=\Theta_{h}$ and $\Theta_{i}=\Theta_{v}$. Nevertheless, as shown in Fig.~\ref{samplecom}, the LiDAR cannot sample perfectly evenly in the inclination/azimuth space. Hence, even if the user manual of the employed LiDAR is followed to set $\Theta_{a}=\Theta_{h}$ and $\Theta_{i}=\Theta_{v}$, the unwanted unobserved regions around or inside the marker's area and the points overlapping issue are still inevitable. Based on the experiments, the points overlapping issue under $\Theta_{a}=\Theta_{h}$ and $\Theta_{i}=\Theta_{v}$ could have an acceptable influence on pose estimation while the unobserved regions impede marker detection. This is the reason why image preprocessing (Refer to Section~\ref{sc}) and the approach to deal with unobserved regions  (Refer to Section~\ref{sd}) are necessary.

\subsection{Image Preprocessing}\label{sc}
\begin{figure*}
	\centering
	\includegraphics[width=17cm]{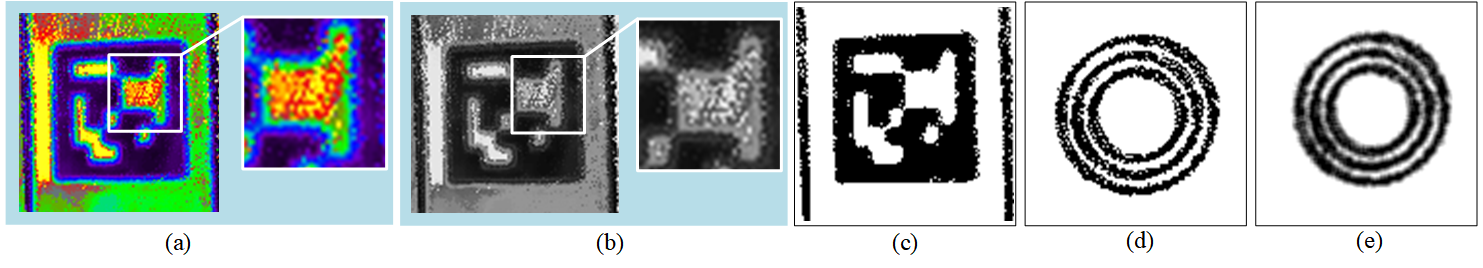}
	\vspace{-0.1in}
	\caption{An illustration of image preprocessing in the system. The input is the raw intensity image (a). The image is first converted to a grayscale image (b). Then it becomes the binary image (c) using naive thresholding. Gaussian Blur is recommended when the embedded VFM system, such as CCTag \cite{cctag}, does not contain it. (d): The intensity image of a CCTag after the naive thresholding while the detector cannot detect the marker in it. (e): The intensity image after applying Gaussian Blur \cite{gb} on (d). Now the marker is detectable.} \label{dis}
\end{figure*}

Fig.~\ref{dis} shows the image preprocessing pipeline before inputting the raw intensity image into the embedded VFM system. Ideally, the intensity values inside the zoomed white-dotted area shown in Fig.~\ref{dis}(a) should be similar whereas these regions have varying low intensity values in the real application. After grayscale conversion, the image becomes Fig.~\ref{dis}(b), where the image noise caused by varying low intensity values still exists. It is found experimentally that if the grayscale intensity image with the quality shown in Fig.~\ref{dis}(b) is inputted into Apriltag 3 \cite{ap3}, the marker detection fails. To remove the image noise shown in Fig.~\ref{dis}(b), following grayscale conversion, binarization is conducted using naive thresholding, which indicates binarization of the entire image with a specified threshold. Fig.~\ref{dis}(c) shows the binarized image. As seen, the noise caused by the varying low intensity values is significantly reduced.

Despite the fact that binarization is a common process in the VFM systems, a mandatory binarization is still added after the grayscale conversion. This is because it is desired that the proposed system have a robust generality when used along with different VFM systems. Detailed explanations are as follows. The binarization processes vary when it comes to different VFM systems as introduced in Section~\ref{stwo}. However, it cannot be guaranteed that the binarization method adopted in the VFM system embedded in our system can handle the image noise shown in Fig.~\ref{dis}(b). Hence, a straightforward but general solution is to add an appropriate and mandatory binarization for the grayscale intensity image. A naive thresholding is adopted due to the following reason. As illustrated in \cite{lt}, the LiDAR measurement is not sensitive to ambient light, which is unlike the camera measurement. When the LiDAR model and the material of the marker are fixed, naive thresholding shows robust denoising performance since the two major elements that affect the intensity are fixed. Some recommended thresholds are provided for different LiDAR models in our program. This threshold is also a custom parameter that the user can adjust according to their LiDAR model and marker material.

Additionally, as seen in Fig.~\ref{dis}(d), random noise could be preserved after the naive thresholding in some cases though this is harmless for VFM systems \cite{ap3,aruco} that adopt Gaussian Blur \cite{gb}. But for systems that do not contain it, such as CCTag \cite{cctag}, Gaussian Blur should be adopted. Otherwise, the marker detection will fail.

\subsection{3D Feature Points Estimation}\label{sd}
The preprocessed image introduced in the previous section is then inputted into the embedded VFM system. Thereafter, the VFM system provides the detection information. In this section, we introduce how to project the 2D feature points given by the detection information back to the 3D space, such that they become 3D feature points expressed in the LiDAR coordinate system. As for the square markers, the feature points refer to the four vertices of the quad. While some of the VFM systems adopt the non-square design, the proposed method has no restriction on the marker shape.

As mentioned in Section~\ref{sa}, the range information is stored for each pixel in the intensity image as well. Thus, a 2D pixel with range information ($\textbf{u}^{r}=[u,v,r]^{T}$) can be projected back to the 3D Cartesian coordinate system by solving the inverse of Eqs. (\ref{pro}-\ref{uv}) to find the corresponding $[x_{l},y_{l},z_{l}]^{T}$. Yet in the real world, it cannot be guaranteed that the feature points of the marker are exactly scanned by the LiDAR. Namely, the detected 2D features in Fig.~\ref{dis}(c) could correspond to the unobserved regions in the raw intensity image. As a matter of fact, this occurs frequently in our experiments, as well as in the dataset provided by \cite{lt}. Hence, when these detected but unscanned features are checked, there will be no range value $r$ returned and it is not feasible to solve the inverse of Eqs. (\ref{pro}-\ref{uv}).
\begin{figure}[htbp]
	\centering
	\includegraphics[width=3.0in]{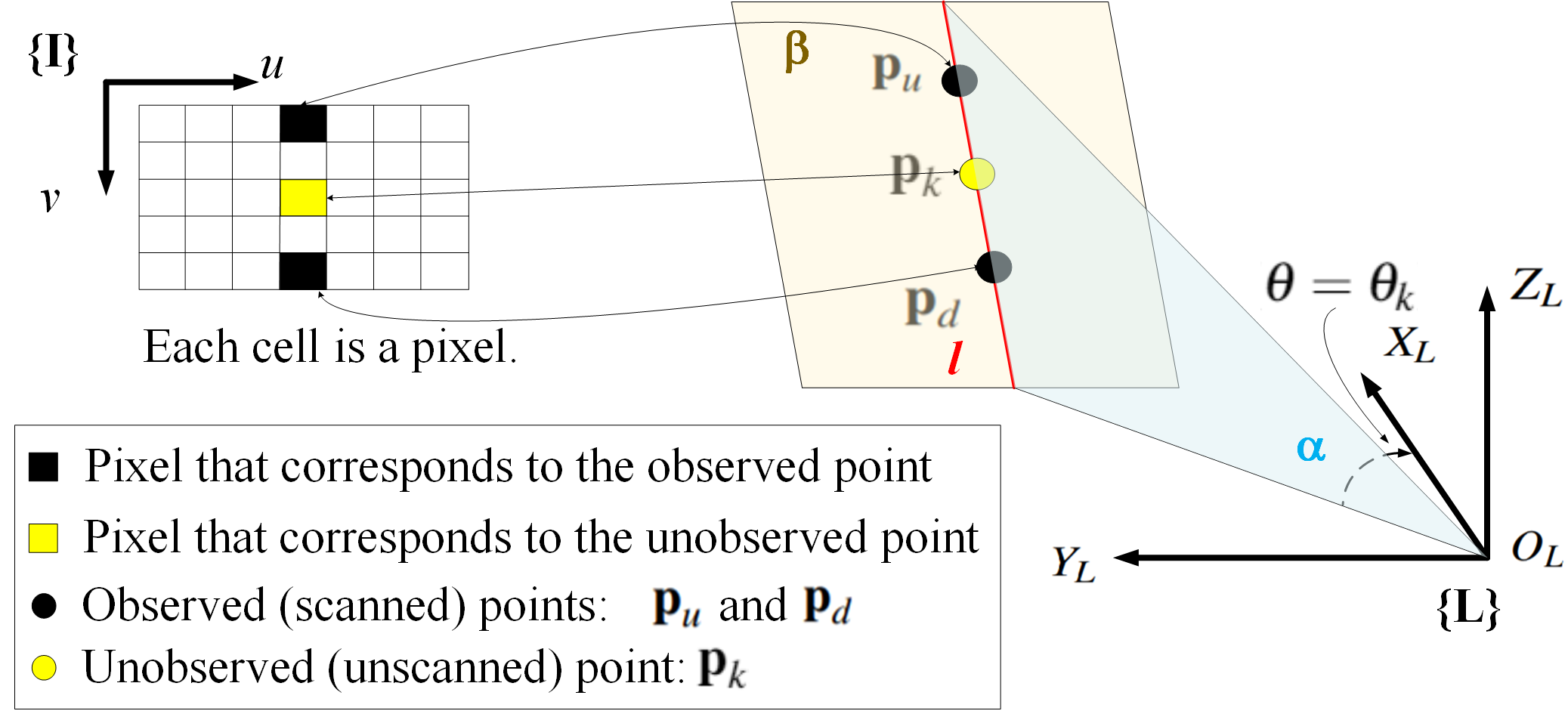}
	\caption{An illustration of the algorithm to estimate the 3D coordinates of a detected but unscanned 2D feature point.}
	\label{collinear}
\end{figure} 

\begin{figure}[thpb]
	\centering
	\includegraphics[width=2.0in]{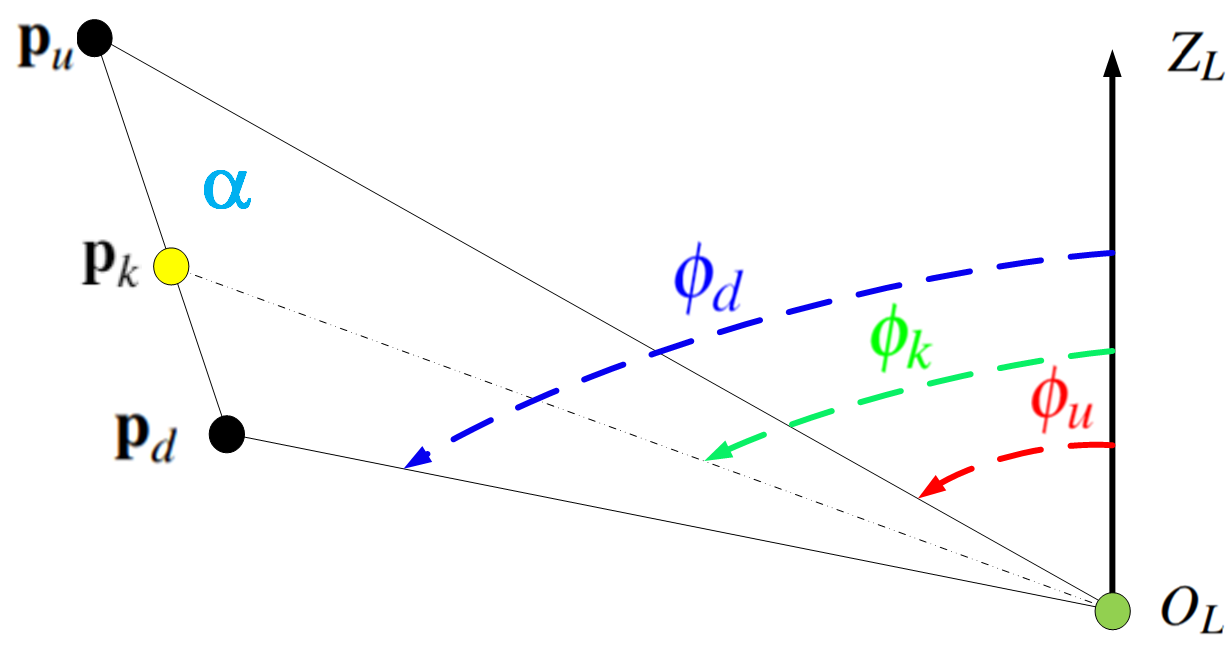}
	\caption{The side view of $\alpha$. $\phi_{d}$, $\phi_{k}$, and $\phi_{u}$ are the inclinations of $\textbf{p}_{d}$, $\textbf{p}_{k}$, and $\textbf{p}_{u}$, respectively. }
	\label{side}
\end{figure}

To resolve this problem, we propose an algorithm as illustrated in Fig.~\ref{collinear}. The algorithm is based on the fact that the markers are planar. Suppose that there is a 2D feature point $\textbf{u}_{k}=[u_{k},v_{k}]^{T}$ that is detected but corresponds to an unobserved region in the raw intensity image. The azimuth $\theta_{k}$ of $\textbf{u}_{k}$ is determined by Eq. (\ref{uv}) through $\theta_{k}=\Theta_{a}(u_{k}-u_{0})$. Define the unknown 3D point corresponding to $\textbf{u}_{k}$ as $\textbf{p}_{k}=[x_{k},y_{k},z_{k}]^{T}$ (the yellow point in Fig.~\ref{collinear}).  Hereafter, suppose that in the same column as $\textbf{u}_{k}$, there is a pair of observed pixels that are symmetric about $\textbf{u}_{k}$, and define their corresponding points as $\textbf{p}_{u}=[x_{u},y_{u},z_{u}]^{T}$ and $\textbf{p}_{d}=[x_{d},y_{d},z_{d}]^{T}$ (the black point in Fig.~\ref{collinear}). As illustrated in Eq. (\ref{uv}), pixels in the same column approximately share the same azimuth. Thus, $\textbf{p}_{u}$, $\textbf{p}_{k}$, and $\textbf{p}_{d}$ are on the same plane, $\alpha$, which is specified by fixing the azimuth ($\theta=\theta_{k}$). After that, define the plane where the marker is located as $\beta$. Obviously, the intersection of $\alpha$ and $\beta$ is a straight line $\textbf{\textit{l}}$. Under the assumption that $\textbf{p}_{u}$ and $\textbf{p}_{d}$  are on the marker plane $\beta$, it can be shown that  $\textbf{p}_{u}$, $\textbf{p}_{k}$, and $\textbf{p}_{d}$ are collinear and they all fall on $\textbf{\textit{l}}$. To give a more explicit introduction to the algorithm, the side view of the plane $\alpha$ is presented in Fig.~\ref{side}.

Undoubtedly, $\textbf{O}_{L}\textbf{p}_{k}$ is the angle bisector of $\angle\textbf{p}_{u}\textbf{O}_{L}\textbf{p}_{d}$ owing to $\phi_{d}-\phi_{k}=\phi_{k}-\phi_{u}$. Hence, in the light of the angle bisector properties, we have 
$\textbf{p}_{k}\textbf{p}_{d}/\textbf{p}_{u}\textbf{p}_{k}=\textbf{O}_{L}\textbf{p}_{d}/\textbf{O}_{L}\textbf{p}_{u}$. Note that $\textbf{O}_{L}\textbf{p}_{d}$ and $\textbf{O}_{L}\textbf{p}_{u}$ are the ranges of $\textbf{p}_{d}$ and $\textbf{p}_{u}$ which can be obtained if they are scanned. Thus, the unknown 3D coordinates of $\textbf{p}_{k}$ are estimated by $\textbf{p}_{k}=\textbf{M}_{1}\textbf{p}_{d}+\textbf{M}_{2}\textbf{p}_{u}$, where $\textbf{M}_{1}=\mathrm{diag}(\frac{\mu}{1+\mu} ,\frac{\mu}{1+\mu} ,\frac{\mu}{1+\mu} ) $ and $\textbf{M}_{2}=\mathrm{diag} ( \frac{1}{1+\mu}, \frac{1}{1+\mu}, \frac{1}{1+\mu})$ with $\mu$ being the ratio $\textbf{O}_{L}\textbf{p}_{d}/\textbf{O}_{L}\textbf{p}_{u}$.

So far we projected the 2D feature points in \textbf{\{I\}}, observed and unobserved by the LiDAR, back to \textbf{\{L\}}.\par
\noindent\textbf{\textit{Remark 2}}: The fundamental assumption of this section is that the 3D points corresponding to nearby pixels of the unobserved feature are still on the marker plane. This is not a harsh requirement since the vertical angular resolution of most LiDAR is less than $2^{\circ}$.

\section{LiDAR Pose Estimation} \label{pose}
The aim of LiDAR pose estimation is to seek the Euclidean transformation, $\textbf{\textit{T}} \in\mathrm{SE(3)}$, from the world coordinate system \textbf{\{W\}} to the LiDAR coordinate system \textbf{\{L\}}. Suppose that $\textbf{\textit{f}} \in \mathbb{R}^{3}$ is the 3D coordinates of a feature point, the operation of $\textbf{\textit{T}} \in\mathrm{SE(3)}$ on $\textbf{\textit{f}} \in \mathbb{R}^{3}$ is $\textbf{\textit{T}} \cdot \textbf{\textit{f}} = \textbf{\textit{R}}\textbf{\textit{f}}+\textbf{\textit{t}}$, where $\textbf{\textit{R}} \in\mathrm{SO(3)}$ is the rotation matrix and $\textbf{\textit{t}}\in\mathbb{R}^{3}$ is the translation vector. Refer to \cite{barfoot} for the basics of $\mathrm{SE(3)}$ and $\mathrm{SO(3)}$.

Thus far two sets of 3D points are obtained. 1) $n$ feature points w.r.t \textbf{\{L\}}, denoted by $\textbf{P}_{L} = \{ \textbf{\textit{f}}_{1}, \ \cdots, \ \textbf{\textit{f}}_{n}\}$, which are given by the 3D feature points estimation introduced in Section~\ref{sd}; 2) $n$ feature points w.r.t \textbf{\{W\}}, denoted by $\textbf{P}_{W}= \{ \textbf{\textit{f}}_{1}^{\prime}, \ \cdots, \ \textbf{\textit{f}}_{n}^{\prime} \}$, which are predefined. Furthermore, the points in $\textbf{P}_{L} $ and $\textbf{P}_{W}$ are matched based on the ID number and vertex index given by the marker detection. Hence, the LiDAR pose estimation can be transformed into finding $\textbf{\textit{R}} \in\mathrm{SO(3)}$ and $\textbf{\textit{t}}\in\mathbb{R}^{3}$ that optimally align $\textbf{P}_{L} $ and $\textbf{P}_{W}$. This is inherently a least square problem:
\begin{equation}	
	\textbf{\textit{R}}^{*},\  \textbf{\textit{t}}^{*}=\underset{\textbf{\textit{R}},\  \textbf{\textit{t}}}{\arg \min } \sum_{j=1}^{n}\left\|\left(\textbf{\textit{R}} \textbf{\textit{f}}_{j}^{\prime}+\textbf{\textit{t}}\right)-\textbf{\textit{f}}_{j}\right\|^{2}\label{least}
\end{equation}

Considering that the point-correspondence between $\textbf{P}_{L} $ and $\textbf{P}_{W}$ is quite reliable thanks to the embedded VFM system, $\textbf{\textit{R}}^{*}$ and $\textbf{\textit{t}}^{*}$ can be calculated in closed form by the Singular Value Decomposition (SVD) method \cite{barfoot,icp}. Moreover, the points in $\textbf{P}_{W}$ are coplanar but not collinear, thus the solution of Eq. (\ref{least}) is also unique and the proof is given in \cite{icp}. This is the reason why the proposed system is free from the multi-solution problem, which is unlike the VFM systems troubled by the rotation ambiguity problem \cite{munoz,munoz2019,ippe,yibo}. This is a superiority of the proposed system over the planar VFM systems \cite{ap3,aruco}.

Remarkably, in the perspective-n-point problem, four correspondences are already enough to estimate the camera pose \cite{ippe}. But when the VFM systems are reviewed, only the four vertices of a marker (or a limited number of points if the marker is non-square) are utilized even though it is definitely feasible to acquire more feature points from the inner coding area of the marker though this increases the complexity of the system. Hence, following this design concept, simply but effectively, only the four vertices of a marker are taken as the feature points. As a result, Table~\ref{tab3} shows that the pose estimation accuracy of the proposed system is slightly inferior to LiDARTag \cite{lt}, which takes thousands of points inside the cluster of the marker as feature points. If one insists on a higher accuracy, it is feasible to use the proposed marker detection information to find the points clustering of the marker in the point cloud and then input the points clustering into the pose estimation block of LiDARTag \cite{lt}.

\section{Experimental Results} \label{exp}
\begin{table*}[htb]
\caption{pose estimation accuracy of the IILFM system at different distances}
\begin{center}
\begin{tabular}{c|c|c|c|c|c|c|c|c|c}
\hline\hline
Distance (m) & Sensor&Number of Scans & System & x (m) &y (m)& z (m)&  roll (deg)& pitch (deg)&  yaw (deg)\\
\hline
    \multirow{7}{*}{$\approx$ 2 m} & \multirow{5}{*}{LiDAR}& -- &Ground Truth &1.604  & -0.158 & 0.612  & -0.015 & -0.090 & -0.014\\ \cline{3-10} 
     & & \multirow{2}{*}{40} &\textbf{IILFM} &1.591  & -0.069 & 0.550  & 0.174  & 4.185  & 1.775\\ \cline{4-10} 
     & & &Error& 0.013  & -0.089 & 0.062  & -0.189 & -4.275 & -1.789\\\cline{3-10} 
       & & \multirow{2}{*}{100} &\textbf{IILFM} &1.601  & -0.164 & 0.629  & -0.438 & 1.930  & -0.873 \\ \cline{4-10} 
       &  & & Error  &0.003  & 0.006  & -0.017 & 0.423  & -2.020 & 0.859 \\
\cline{2-10} 
  & \multirow{3}{*}{Camera} & \multirow{3}{*}{--} &Ground Truth&1.615  & -0.086 & 0.167  & 0.020 & 1.879  & 0.275 \\ \cline{4-10}
   & & & Apriltag 3  &1.631  & -0.120  & 0.183 & -0.787  &5.045 & -0.969 \\ \cline{4-10}
       & & & Error  &-0.016  &0.034  & -0.016& 0.807 &-3.166 & 1.244 \\
\hline
    \multirow{7}{*}{$\approx$ 3 m} &\multirow{5}{*}{LiDAR}&-- &Ground Truth & 0.612  & -0.152 & 0.624  & -0.007 & -0.103 & -0.031 \\ \cline{3-10} 
     & &\multirow{2}{*}{40} &\textbf{IILFM}  &0.583  & -0.133 & 0.625  & -0.733 & 2.821  & 1.016\\ \cline{4-10} 
     & &  &Error& 0.029  & -0.019 & -0.001 & 0.726  & -2.924 & -1.047\\\cline{3-10} 
    &   &\multirow{2}{*}{100}&\textbf{IILFM} &0.586  & -0.131 & 0.717  & -0.937 & 1.079  & 1.080 \\ \cline{4-10} 
   &    & &Error  &0.026  & -0.021 & -0.093 & 0.930  & -1.182 & -1.111  \\
\cline{2-10}

 & \multirow{3}{*}{Camera} & \multirow{3}{*}{--} &Ground Truth&0.599  & -0.051 & 0.184  & 0.073 & 0.360  & 0.321 \\ \cline{4-10} 
       & & & Apriltag 3  &0.527  & 0.073  & 0.228 & 1.442  & -7.603 & 3.225 \\ \cline{4-10}
         & & & Error  &0.072  &-0.124  &-1.369 & 0.423  & 7.963 & -2.904 \\
\hline
    \multirow{7}{*}{$\approx$ 4 m} &\multirow{5}{*}{LiDAR}&-- &Ground Truth  &-0.400 & -0.163 & 0.632  & -0.005 & -0.088 & -0.015\\ \cline{3-10} 
     &&\multirow{2}{*}{40} &\textbf{IILFM}  &-0.422 & -0.311 & 0.597  & -1.437 & 2.288  & -2.138 \\ \cline{4-10} 
   &  & &Error& 0.022  & 0.148  & 0.035  & 1.432  & -2.376 & 2.123 \\\cline{3-10} 
    &   &\multirow{2}{*}{100}&\textbf{IILFM}  & -0.424 & -0.187 & 0.739  & -0.347 & 0.369  & -0.406 \\ \cline{4-10} 
      & & &Error   & 0.024  & 0.024  & -0.107 & 0.342  & -0.457 & 0.391 \\
    \cline{2-10}  
       & \multirow{3}{*}{Camera} & \multirow{3}{*}{--} &Ground Truth&-0.399  & -0.091 & 0.199  & 0.072 & 1.395  & 0.845 \\ \cline{4-10} 
       & & & Apriltag 3  &-0.341  & 0.316  & 0.432 & 1.505  & -7.897 & 14.279 \\ \cline{4-10}
          & & & Error  &-0.058 & 	-0.407  &	-0.233 & -1.433  & 9.292 &-13.434\\
\hline\hline
\end{tabular}
\label{tab1}
\end{center}
\end{table*}

\subsection{Qualitative Evaluation}
\begin{figure}[htpb]
	\centering
	\includegraphics[width=2.8in]{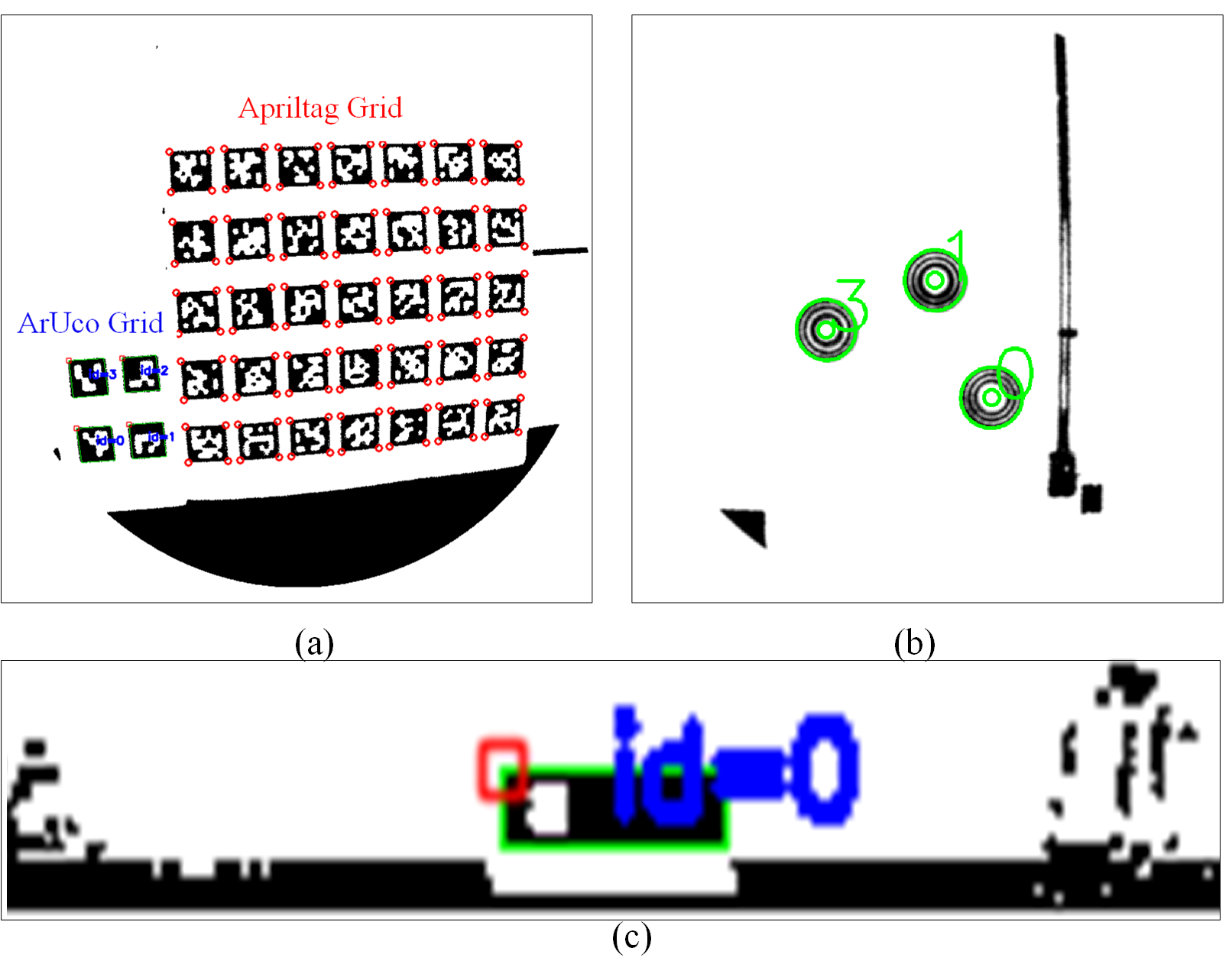}
	\vspace{-0.1in}
	\caption{Marker detection results on the preprocessed intensity images.\textbf{ (a)} corresponds to the scenario shown in Fig.~\ref{flex}(a): a Livox Mid-40 is scanning an Apriltag grid and an ArUco grid. All the markers, 35 Apriltags (family: tag36h11, marker size: 17.2 cm$\times$17.2 cm), and 4 ArUcos (family: 4$\times$4, marker size: 16 cm$\times$16 cm), are detected. \textbf{(b)} corresponds to Fig.~\ref{flex}(b): a Livox Mid-40 is scanning three CCTag (family: 3 rings, marker radius: 7.2 cm) attached to the wall. All the markers are detected. \textbf{(c)} corresponds to Fig.~\ref{flex}(c): a VLP-16 is scanning an ArUco (family: original, maker size: 40 cm$\times$40 cm). The marker is detected.} \label{result}
\end{figure}

To qualitatively demonstrate the flexibility of the proposed IILFM system, Fig.~\ref{result} is presented. The usage of the IILFM system is as convenient as the VFM systems. In particular, the user can place the Letter size markers \cite{ap3,aruco} densely to compose a marker grid, as shown in Fig.~\ref{result}(a), as well as attach some non-square markers \cite{cctag} to the wall freely, as shown in Fig.~\ref{result}(b). In summary, there is no spatial restriction on marker placement and this is the most important virtue of our system. It should be noted that if the angular resolution of the LiDAR is relatively large, large marker size and simpler marker pattern are recommended for the sake of the intensity image quality. For instance, the vertical angular resolution of the VLP-16 is 1.33$^{\circ}$, thus a 40 cm$\times$40 cm original ArUco \cite{aruco} is adopted, as shown in Fig.~\ref{result}(c).

\subsection{Quantitative Evaluation}
\begin{figure}[htpb]
	\centering
	\includegraphics[width=2.8in]{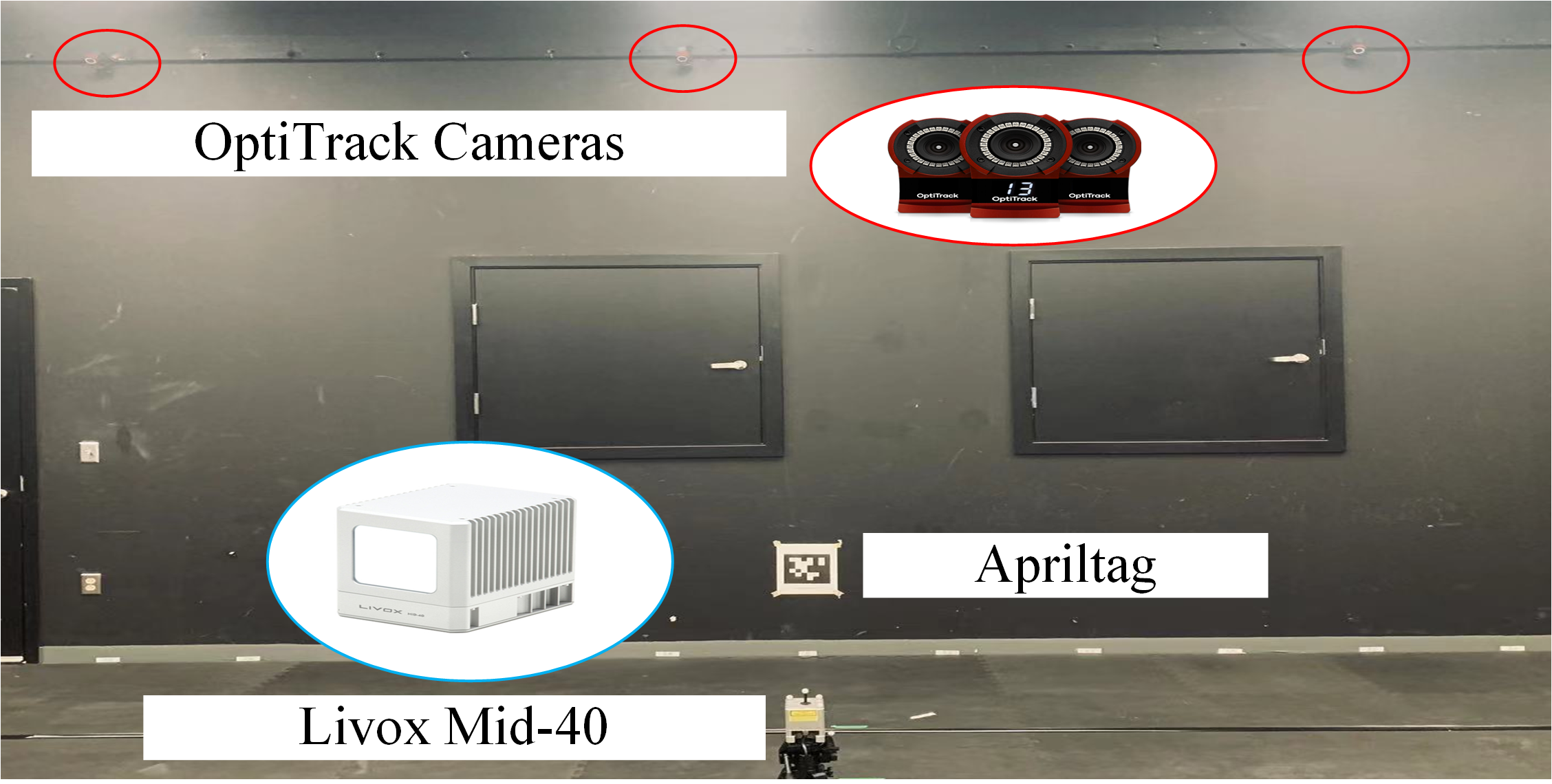}
	\vspace{-0.1in}	
	\caption{An illustration of the experimental setup. The MoCap system is composed of 16 OptiTrack cameras and provides the 6-DOF pose information of the predefined rigid body at 100 Hz.} \label{setup}
\end{figure}

To further verify the pose estimation precision of the IILFM system, we compare the pose estimation result given by our system with the ground truth provided by the OptiTrack Motion Capture (MoCap) system (See Fig.~\ref{setup}). Again, we adopt the low-cost solid-state LiDAR, Livox Mid-40, to scan a marker pasted on the wall. The marker (ID = 0), with the size of 17.2 cm$\times$17.2 cm, belongs to the \textit{tag36h11} family of Apriltag 3 \cite{ap3} and is printed on Letter paper. The location of the marker's center is at (3.620, 0.00, 0.485) m w.r.t \textbf{\{W\}}.

Firstly, the orientation of the LiDAR is fixed, with the roll, pitch, and yaw angles approximately being zeros. Only the distance from the marker's plane to the LiDAR's $O_{L}Y_{L}\mbox{-}O_{L}Z_{L}$ plane, see Table~\ref{tab1}, is changed. Then, to compare the conventional VFM system with the proposed system, we test Apriltag 3 \cite{ap3} with a camera (Omnivision OV7251) under the same experimental setup. Three issues are illustrated in Table~\ref{tab1}: 1) When more LiDAR scans are utilized, the pose estimation precision is slightly boosted. The reason behind it is that more LiDAR scans indicate a higher coverage percentage in the FoV \cite{loam}, which implies better intensity image quality. 2) The IILFM system shows comparable precision as the VFM system. 3) Unlike the VFM system \cite{wang,olson,ap3}, the pose estimation precision does not degrade evidently as the distance increases.

Thereafter, the distance from the marker to the LiDAR is set as 2 meters and only the rotation of the LiDAR is adjusted (See Table~\ref{tab2} where the number of scans = 40). By comparing the errors in Table~\ref{tab2} and Table~\ref{tab1}, it is seen that when the plane of the LiDAR is angled towards the marker plane, the pose estimation performance of the IILFM system is as good as that when the LiDAR is perpendicular to the marker.

\begin{table}[htb]
\caption{pose estimation accuracy of the IILFM system with different Euler angles}
\begin{center}
\begin{tabular}{c|c|c|c|c}
\hline\hline
Setup & Term & Ground Truth &\textbf{IILFM}&Error\\ \hline
 \multirow{6}{*}{ pitch $\approx -15^{\circ}$} &x (m) &1.629   & 1.661   & -0.032  \\  \cline{2-5} 
 &y (m) &-0.066  & -0.011  & -0.055  \\  \cline{2-5} 
&z (m) &0.618   & 0.699   & -0.081  \\  \cline{2-5} 
 &roll (deg) &-2.673  & -4.936  & 2.263  \\  \cline{2-5} 
  &pitch (deg) &\textbf{-15.060} & -21.145 & 6.085 \\  \cline{2-5} 
   &yaw (deg) &-0.171  & 0.546   & -0.717 \\  \hline
 \multirow{6}{*}{ pitch $\approx 15^{\circ}$} &x (m) &1.684   & 1.622   & 0.062    \\  \cline{2-5} 
 &y (m) &-0.063  & -0.088  & 0.026  \\  \cline{2-5} 
&z (m) &0.590   & 0.660   & -0.070  \\  \cline{2-5} 
 &roll (deg) &3.657   & 0.900   & 2.757 \\  \cline{2-5} 
  &pitch (deg) &\textbf{14.849}  & 10.961  & 3.888 \\  \cline{2-5} 
   &yaw (deg) &1.136   & -1.921  & 3.057 \\  \hline
    \multirow{6}{*}{ yaw $\approx -15^{\circ}$} &x (m) &1.638   & 1.612   & 0.027   \\  \cline{2-5} 
 &y (m) &-0.618  & -0.237  & -0.381 \\  \cline{2-5} 
&z (m) &0.610   & 0.585   & 0.025  \\  \cline{2-5} 
 &roll (deg) &-0.009  & -0.792  & 0.783 \\  \cline{2-5} 
  &pitch (deg) &-0.662  & -2.088  & 1.426\\  \cline{2-5} 
   &yaw (deg) &\textbf{-15.387} & -17.937 & 2.550 \\  \hline
    \multirow{6}{*}{ yaw $\approx 15^{\circ}$} &x (m) &1.627   & 1.632   & -0.005  \\  \cline{2-5} 
 &y (m) &-0.194  & -0.146  & -0.048  \\  \cline{2-5} 
&z (m) &0.609   & 0.553   & 0.057   \\  \cline{2-5} 
 &roll (deg) &0.490   & -1.994  & 2.484 \\  \cline{2-5} 
  &pitch (deg) &-0.388  & -0.359  & -0.028 \\  \cline{2-5} 
   &yaw (deg) &\textbf{14.924}  & 10.678  & 4.246 \\  \hline\hline
\end{tabular}
\label{tab2}
\end{center}
\end{table}

To validate the pose estimation accuracy of our system on the mechanical LiDAR as well as to compare the proposed system with the state-of-the-art LiDAR fiducial marker system, LiDARTag \cite{lt}, which is also the only existing fiducial marker system for the LiDAR as far as we know, Table~\ref{tab3} is presented. Specifically, we use the rosbag (ccw\_10m.bag) provided by \cite{lt} as the benchmark on which we conduct the comparison. ccw\_10m.bag records the raw data of a 32-Beam Velodyne ULTRA Puck LiDAR scanning a 1.22 m$\times$1.22 m Apriltag (tag16h6), from a distance of 10 meters while the relative angle between the LiDAR plane and the marker is around 45$^{\circ}$. Only the vertices estimation is compared on account that \cite{lt} solely provides the ground truth of the vertices and the vertices estimation is a follow-up process after the pose estimation in the LiDARTag system \cite{lt,lt2}. Hence, the vertices estimation can also reflect the pose estimation precision. Table~\ref{tab3} illustrates that the IILFM system is slightly inferior to LiDARTag in terms of accuracy, whereases as mentioned previously, the superiority of the IILFM system is the flexibility and convenience. Marker detection in the scenarios shown in Fig.~\ref{flex} and Fig.~\ref{setup} is infeasible for the LiDARTag system since the marker placement does not satisfy the requirement of LiDARTag, and in addition, the current version of LiDARTag does not support any non-square marker, such as CCTag \cite{cctag}.

\begin{table}[htbp]
\caption{Comparison of the IILFM system and LiDARTag}
\begin{center}
\begin{tabular}{c|c|c|c|c|c}
\hline\hline
System & Vertex & x (m) &y (m)& z (m) & Error (m)\\ \hline
 \multirow{4}{*}{Ground Truth } &1 &9.739 & 0.758 & -0.161 & --  \\  \cline{2-6} 
 &2 &9.940 & 0.072 & -0.732 & -- \\  \cline{2-6} 
&3 &10.272 & -0.414 & -0.032 & -- \\  \cline{2-6} 
\cite{lt} &4 &10.072 & 0.271 & 0.539 & --\\  \hline
 \multirow{4}{*}{LiDARTag } &1 & 9.736 & 0.762 & -0.174 & 0.015\\  \cline{2-6} 
 &2 &9.944 & 0.066 & -0.730 & 0.009 \\  \cline{2-6} 
&3 &10.271 & -0.405 & -0.016 & 0.019 \\  \cline{2-6} 
\cite{lt} &4 & 10.063 & 0.291 & 0.539 & 0.022\\  \hline
 \multirow{4}{*}{IILFM } &1 &9.728 & 0.766 & -0.184 & 0.013\\  \cline{2-6} 
 &2 & 9.963 & 0.052 & -0.705 & 0.033 \\  \cline{2-6} 
&3 &10.307 & -0.432 & -0.016 & 0.044 \\  \cline{2-6} 
&4 &10.045 & 0.263 & 0.564 & 0.041 \\  \hline\hline
\end{tabular}
\label{tab3}
\end{center}
\end{table}

\subsection{Computational Time Analysis}
A computational time analysis is conducted on a desktop with Intel Xeon W-1290P CPU. The LiDARTag \cite{lt} runs at around 100 Hz on it. The time consumption of the marker detection of our system, which includes intensity image generation, preprocessing, 2D features detection, and computation of 3D features, mainly depends on the size of the intensity image and the embedded VFM system. Suppose that the embedded VFM system is Apriltag 3 \cite{ap3}, the marker detection takes around 7 ms (approx 143 Hz) in the case shown in Fig.~\ref{result}(c) where $\Theta_{a}=0.3^{\circ}$, $\Theta_{i}=1.33^{\circ}$, and intensity image size = $1201\times27$. For the case shown in  Fig.~\ref{result}(a), the marker detection takes around 25 ms (approx 40 Hz), where intensity  $\Theta_{a}=\Theta_{i}=0.05^{\circ}$ and image size = $771\times591$. The time consumption of the following pose estimation process is around 1.8 $\upmu$s as the closed-form solution can be obtained directly through SVD \cite{icp} (Refer to Section~\ref{pose}).

\subsection{Limitations Analysis}
There are some limitations on the application of the LiDAR fiducial marker systems. First, the distance from the object (marker) to the LiDAR must exceed the minimum detectable range. This limitation is caused by the hardware attributes and it affects all the LiDAR applications not only the LiDAR fiducial marker systems. Secondly, to utilize the solid-state LiDAR, it is required to wait for the growth of the scanned area inside the FoV to obtain a relatively dense point cloud. Again this is a limitation caused by the hardware attributes. In contrast, the mechanical LiDAR only requires one LiDAR scan to work, however, a larger and simpler marker is recommended if the angular resolution is large.  Finally, the false positives are occasionally found in the experiments while the issue of wrong ID detection (not exactly the same as the false positive but similar) is also reported in LiDARTag \cite{lt}. For the proposed system, the false positive rate is mainly determined by the embedded VFM system, thus novel VFM systems with lower false positive rate, such as \cite{ap3,aruco}, are preferred.

\section{Conclusions and Future Work} \label{con}
In this work, a novel intensity image-based LiDAR fiducial marker is proposed. The main novelty of the system is its flexibility and convenient application. Namely, the users can place the markers the same way as placing the conventional visual fiducial markers without worrying about any spatial restriction on marker placement. Following the method introduced in this work, the popular visual fiducial marker systems, such as Apriltag, ArUco, CCTag, and newly proposed ones in the future can be easily integrated into the system, thus the proposed system has no restriction on marker shape. Experimental results demonstrate that the proposed system has the similar accuracy as the visual fiducial marker system in terms of pose estimation.

As for future work, motion compensation is a vital direction since it is assumed that the unstructured point cloud inputted to the system is already motion compensated. Furthermore, the intensity image quality could be unideal due to the over-sparsity of the point cloud, which hinders the marker detection. Hence, migrating the deep-learning-based 3D point cloud completion method \cite{grnet} into the system is also an interesting topic. In addition, we are developing a LiDAR fiducial marker-based SLAM framework. 

%
%
\section*{ACKNOWLEDGMENT}
The authors would like to thank Yicong Fu, Hassan Alkomy, and Brian Lynch for constructive discussions, Jiunn-Kai Huang for providing his experimental results. 
%
%
%
%

\bibliographystyle{IEEEtran} 
\bibliography{reference}

\end{document}